# MDD-LLM: Towards Accuracy Large Language Models for Major Depressive Disorder Diagnosis

Yuyang Sha[1], Hongxin Pan[1], Wei Xu[1], Weiyu Meng[1], Gang Luo[1], Xinyu Du[2], Xiaobing Zhai[1], Henry H. Y. Tong[1], Caijuan Shi[3], Kefeng Li[1*]

[1]Center for Artificial Intelligence Driven Drug Discovery, Faculty of Applied Sciences, Macao Polytechnic University, Macau, 999708, Macau SR
[2]Department of Dentistry, Botou Hospital, Cangzhou, 062150, HeHei, China
[3]College of Artificial Intelligence, North China University of Science and Technology, TangShan, 063210, HeBei, China

\* Corresponding author Kefeng Li, PhD.
\* Email: kefengl@mpu.edu.mo

## Abstract

Major depressive disorder (MDD) impacts more than 300 million people worldwide, highlighting a significant public health issue. However, the uneven distribution of medical resources and the complexity of diagnostic methods have resulted in inadequate attention to this disorder in numerous countries and regions. This paper introduces a high-performance MDD diagnosis tool named MDD-LLM, an AI-driven framework that utilizes fine-tuned large language models (LLMs) and extensive real-world samples to tackle challenges in MDD diagnosis. Therefore, we select 274,348 individual information from the UK Biobank cohort to train and evaluate the proposed method. Specifically, we select 274,348 individual records from the UK Biobank cohort and design a tabular data transformation method to create a large corpus for training and evaluating the proposed approach. To illustrate the advantages of MDD-LLM, we perform comprehensive experiments and provide several comparative analyses against existing model-based solutions across multiple evaluation metrics. Experimental results show that MDD-LLM (70B) achieves an accuracy of 0.8378 and an AUC of 0.8919 (95% CI: 0.8799 - 0.9040), significantly outperforming existing machine learning and deep learning frameworks for MDD diagnosis. Given the limited exploration of LLMs in MDD diagnosis, we examine numerous factors that may influence the performance of our proposed method, such as tabular data transformation techniques and different fine-tuning strategies. Furthermore, we also analyze the model's interpretability, requiring the MDD-LLM to explain its predictions and provide corresponding reasons. This paper investigates the application of LLMs and large-scale training data for diagnosing MDD. The findings indicate that LLMs-driven schemes offer significant potential for accuracy, robustness, and interpretability in MDD diagnosis compared to traditional model-based solutions.

**Keywords:** major depressive disorder, large language models, artificial intelligence, model fine-tuning, medical data processing.

## 1. Introduction

Major depressive disorder (MDD) is a serious mental health condition that typically exerts a detrimental influence on an individual's daily life, occupational performance, and physical well-being [1]. The World Health Organization (WHO) estimates that over 300 million people around the world are affected by MDD [2]. Research indicates that over 70% of suicide or self-harm behaviors are associated with MDD [3]. Currently, MDD is recognized as the second largest health issue. Studies [4] forecast that by around 2030, MDD is expected to become the leading health condition worldwide. Despite the substantial resources allocated to researching the causes, diagnosis, and treatment of MDD, progress in these areas remains below expectations. In certain regions, individuals with MDD may face cultural and economic barriers that limit their access to appropriate treatment [5]. The development of diagnostic technologies for MDD is a key research topic, underscoring its considerable academic and practical value. Currently, most diagnostic methodologies for MDD rely on standardized scales, such as the Patient Health Questionnaire-9 (PHQ-9) and the Hamilton Depression Rating Scale (HDRS) [6]. However, these diagnostic approaches require active cooperation from patients, and the results can be easily influenced by diverse external factors. Therefore, some studies have sought to construct scale-independent diagnostic schemes for MDD. For example, Watts et al. [7] developed a machine-learning approach to predict MDD risk factors using electroencephalography. Similarly, Pan et al. [8] utilized metabolomic networks combined with advanced deep-learning techniques to identify signatures for classifying MDD. Additionally, several researchers [9–12] are exploring more effective diagnostic solutions for MDD identification by leveraging multimodal information, including MRI, molecular techniques, and various signals. While these approaches aim to create efficient diagnostic frameworks for MDD from multiple perspectives, they still encounter various challenges. For example, most model-based MDD diagnostic frameworks are data-driven, meaning their effectiveness heavily relies on the quantity and quality of the provided training data. However, obtaining sufficient data can often be difficult in the medical field, particularly for certain rare diseases [13].

While the precise pathogenesis of MDD remains unclear, extensive studies have examined the interaction between biological and social factors associated with MDD. For example, Blundell et al. [14] established a correlation between Body Mass Index (BMI) and MDD. By analyzing a substantial dataset from retrospective cohort data, the researchers demonstrated that individuals with obesity frequently exhibit a heightened risk of developing MDD. Darling et al. [15] examined the link between work-related pressure and MDD, showing that prolonged and intense work conditions can increase the likelihood of depression. In addition, multiple factors [16, 17] such as income, occupation, education level, and marital status, have been found to be associated with the prevalence of MDD. However, current model-based MDD diagnostic schemes struggle to utilize this prior knowledge effectively. The continuous advancement of artificial intelligence technology has promoted the rapid development of foundational large language models (LLMs), which exhibit considerable potential in many natural language processing tasks [18, 19]. Therefore, researchers are exploring the application of LLMs to the medical field and have achieved promising results in several tasks, such as medical report generation [20], health diagnostic chatbots [21], and clinical AI agents [22]. Existing LLMs are trained on large amounts of samples, allowing them to understand and generate natural language. For instance, LLMs can comprehend the harmful impacts of diabetes on the human body and recognize the normal ranges for blood pressure [23]. In contrast, traditional model-based solutions [24–26] such as CNN, LSTM, and XGBoost usually lack this capability. Moreover, LLMs offer several advantages when handling medical tasks. Data cleaning is critical in the medical field, often requiring complex processing of outliers and missing values. Fortunately, LLMs exhibit a commendable tolerance for these issues, making them particularly suitable for medical applications [27]. Furthermore, LLMs-based solutions predominantly rely on conversational interactions, facilitating their ability to comprehend the input information and effectively elucidate the results they generate. As a result, medical models powered by LLMs offer stronger interpretability and greater potential for subsequent development.

In this paper, we present a high-performance tool for MDD diagnosis named MDD-LLM. This scheme employs classical LLMs in conjunction with parameter-efficient fine-tuning techniques to enhance diagnostic accuracy and efficiency. Generally, fine-tuning LLMs demands a considerable volume of high-quality training samples. To address this challenge, we select 274,348 individual records from the UK Biobank cohort [28] and design a tabular data transformation method to create a large corpus for training and evaluating the proposed approach. During the fine-tuning stage, the LLMs need to learn from the input information, determine whether the individual suffers from MDD, and provide the probability of that determination. To demonstrate the effectiveness of our proposed method, we also implement several classification models for MDD diagnosis based on the same dataset, including ResNet1D [29], Multi-layer Perceptron (MLP), XGBoost, and Random Forest (RF). Experimental results indicate that the proposed MDD-LLM (70B) achieves an accuracy of 0.8378 and an AUC of

0.8919 (95% CI: 0.8799 - 0.9040), significantly outperforming existing machine learning and deep learning frameworks for MDD diagnosis. Additionally, we investigate numerous factors that may influence the performance of MDD-LLM, such as tabular data transformation techniques, different fine-tuning strategies, and the effectiveness of handling missing features. Furthermore, we design experiments to assess the interpretability of the model and develop a graphical user interface to enhance the usability of our proposed MDD-LLM.

## 2. Materials and Methods

**2.1 Dataset.**

The training process for LLMs typically involves two stages: pre-training and supervised fine-tuning (SFT). This article focuses on the SFT phase, emphasizing the importance of both the quantity and quality of the training data. Data collection and organization within the medical field have consistently posed significant challenges. Data collection and organization in medical-related tasks often require addressing many challenges. Fortunately, the UK Biobank cohort [28] provides a large amount of medical data, particularly regarding MDD diagnosis, which serves as the primary resource for experimentation and analysis in this article.

The UK Biobank is a large-scale prospective cohort study that recruited over 500,000 individuals (aged 40-69) between 2006 and 2010. This project was conducted across 22 assessment centers in England, Wales, and Scotland. Participants' health information, hospital records, urine and blood biomarkers, and genetic data were gathered through touchscreen questionnaires and verbal interviews. All participants provided electronic informed consent when they were recruited. In this study, the phenotype for MDD was defined using the International Statistical Classification of Diseases and Related Health Problems, 10th Revision (ICD-10) code F32, corresponding to a depressive episode [30]. This definition also encompasses self-reported conditions documented during the assessment, as noted in data fields 20433 and 20434. Our study comprises 274,348 participants, including 12,715 diagnosed with MDD and 261,633 control subjects. The baseline characteristics of UK Biobank are shown in Table 1.

Table 1. The baseline characteristics of participating population.

| Characteristics | HC ($n$ = 261,633) | MDD ($n$ = 12,715) | $P$ value |
|---|---|---|---|
| Age | 61 (53 - 66) | 57 (50 - 63) | < 0.001 |
| BMI | 26.96 (24.35 - 30.12) | 26.84 (24.03 - 30.23) | 0.051 |
| HDL cholesterol | 0.13 (0.61 - 0.33) | 0.13 (0.62 - 0.33) | 0.124 |
| Clinical LDL Cholesterol | -0.05 (0.64 - 0.70) | -0.05 (0.63 - 0.70) | 0.843 |
| Total Cholesterol | -0.03 (0.64 - 0.68) | -0.04 (0.62 - 0.67) | 0.664 |
| Triglycerides | -0.20 (0.42 - 0.62) | -0.20 (0.41 - 0.61) | 0.684 |
| Sleep Duration, median | 7 (6 - 8) | 7 (6 - 8) | < 0.001 |
| Sex (%) | | | < 0.001 |
|   Female | 140,980 (53.88) | 8,081 (63.55) | |
|   Male | 120,653 (46.12) | 4,634 (36.45) | |
| Sleeplessness (%) | | | < 0.001 |
|   Usually | 72,840 (27.84) | 4439 (34.91) | |
|   Sometimes | 124,256 (47.49) | 5940 (46.71) | |
|   Never | 63,741 (24.36) | 2309 (18.16) | |
|   Missing Value | 796 (0.31) | 27 (0.22) | |
| Alcohol Frequency (%) | | | < 0.001 |
|   Usually | 52,847 (20.20) | 2,417 (19.01) | |
|   Sometimes | 187,034 (71.48) | 9,189 (72.30) | |
|   Never | 20,961 (8.02) | 1,067 (8.39) | |
|   Missing Value | 791 (0.30) | 42 (0.30) | |
| Self-harmed Action (%) | | | < 0.001 |
|   Yes | 3,189 (1.21) | 597 (4.69) | |
|   No | 74,140 (28.33) | 7,831 (61.58) | |
|   Not Answer | 211 (0.08) | 31 (0.24) | |
|   Missing Value | 184,083 (70.38) | 4,256 (33.49) | |
| Education (%) | | | 0.292 |
|   Low | 44,503 (17.01) | 2,189 (17.22) | |
|   Intermediate | 128,017 (48.93) | 6,236 (49.05) | |
|   High | 83,827 (32.04) | 4,039 (31.69) | |
|   Missing Value | 5316 (2.03) | 260 (2.04) | |
| Income (%) | | | 0.608 |
|   Low | 52,023 (19.88) | 2,558 (20.11) | |
|   Medium | 159,755 (61.07) | 7773 (61.46) | |
|   High | 11,273 (4.30) | 567 (4.14) | |
|   Missing Value | 38,582 (14.75) | 1,817 (14.29) | |
| Employ Status (%) | | | 0.652 |
|   Employed | 149,026 (56.96) | 7,275 (57.08) | |

| | | |
|---|---|---|
| Not Employed | 108,525 (41.48) | 5,238 (41.20) |
| Other Status | 2,616 (1.00) | 139 (1.11) |
| Missing Value | 1,466 (0.56) | 81 (0.61) |

* Abbreviations: BMI, body mass index, calculated as weight in kilograms divided by height in meters squared.

Education: low (no relevant qualifications); intermediate (A levels, AS levels, or equivalent; O levels, GCSEs, or equivalent; CSEs or equivalent; NVQ or HND or HNC or equivalent; and other professional qualifications); high (college or university degree).

Income is defined by average total household income before tax: low (less than £18,000); medium (£18,000 to £30,999; £31,000 to £51,999 and £52,000 to £100,000); high (greater than £100,000).

Current employment status: employed (in paid employment or self-employed and full or part-time student); not employed (retired; unable to work because of sickness or disability; unemployed and looking after home and/or family); other status (doing unpaid or voluntary work and none of the above).

## 2.2 Tabular Data Processing

Similar to other medical datasets, the majority of data within the UK Biobank is organized in a tabular format. However, current LLMs struggle to process tabular data directly. Therefore, it is necessary to convert the tabular information into instruction prompts that LLMs can understand. We propose three tabular data conversation methods, including Text Template, List Template, and GPT Generation Template. Specifically, the List Template employs a straightforward format for data conversion, such as "{Feature Name} is {Value}." The Text Template is more complex than the List Template, which requires creating several short sentences based on different feature names. The GPT Generation Template utilizes the ChatGPT API, allowing the online ChatGPT model to transform input tabular data into descriptive text. The details of these conversation methods are shown in Appendix S1. For instance, we utilize the data presented in Table 2 as a case study to conduct data transformation by Text Template.

Table 2. An example for tabular data conversation.

| Age | BMI | Sleepless | Sleep Times | Drink | Self-harmed | Employment |
|---|---|---|---|---|---|---|
| 60 | 24.5018 | Sometime | 6h | 3/week | Never | Paid |
| Work Times | Education | Income | HDLC | CLDLC | TG | TC |
| 38h/week | O Level | 45000 | 2.075 | 2.6077 | 1.334 | 4.7848 |

The generated prompt sample by Text Template method are shown as follow:

"*instruction*": "Predict if a patient has the major depressive disorder? Yes or no? Please answer with only yes or no and do not give any extra information."

"*input*": "Age is 60, sex is female, body mass index (BMI) is 24.5018 kg/m^2, sometimes sleeplessness, sleep time is 6 hours, drink alcohol three times a week, never self-harmed, the employment status is in paid employment, the income is 45000 pound, work 38 hours per week, the education is o levels, not has long-standing illness, the hdl cholesterol is 2.075 mmol/l, the clinical ldl cholesterol is 2.6077 mmol/l, the triglycerides is 1.334 mmol/l, the total cholesterol is 4.7848 mmol/l",

"*output*": "Yes"

In this context, "*instruction*" refers to the provided system prompt, "*input*" denotes the transformed information from the tabular data, and "*output*" indicates the desired result we expect from the model. In particular, it is essential for the model to not only provide the corresponding answer but also to deliver the associated probability for that answer. This probability functions as a risk score and can be evaluated through metrics to assess the model's discriminatory power. Additionally, several transformations of tabular data may significantly impact the outcomes of our proposed MDD-LLM. In the experimental section, we present a comprehensive discussion on the effectiveness of these different transformation methods.

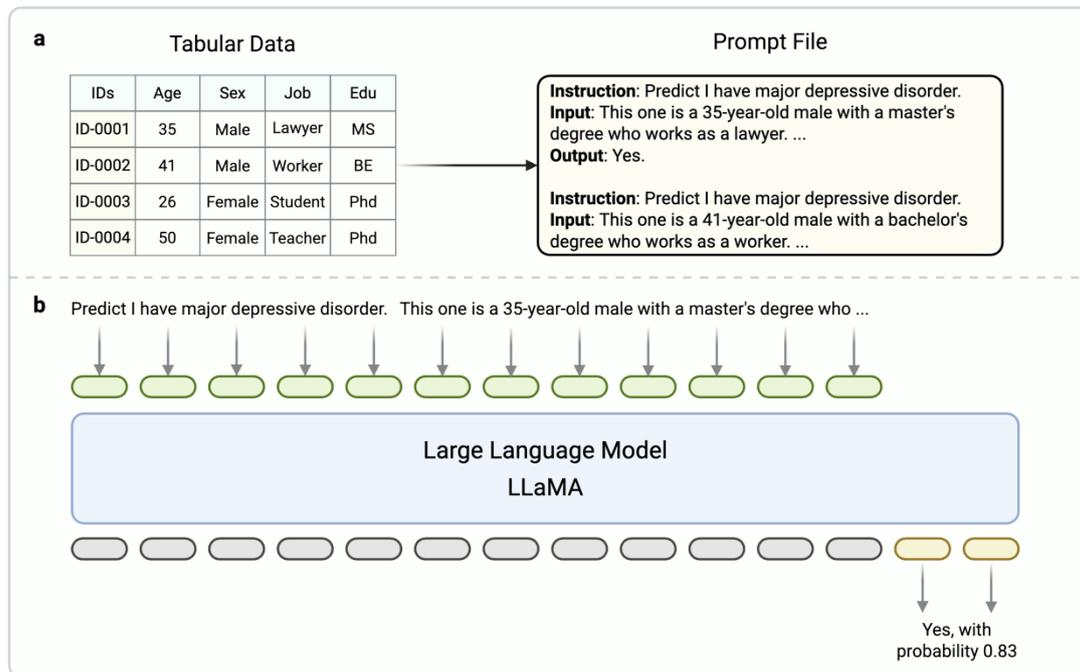

Figure 1. The structure of our proposed method. Figure 1a demonstrate the tabular data transformation process. Figure 1b defines the overview of MDD-LLM. The generated prompt information is tokenized and embedding into the token embedding space. Then the proposed MDD-LLM should give the prediction result and probability by the inputs.

## 2.3 Proposed Method

To achieve an optimal solution for diagnosing MDD, it is necessary to fine-tune the appropriate LLMs using high-quality data to enhance the model's capabilities in psychology. The task involved in this paper can be simplified to a special classification problem. As a critical component, the LLM should understand the transformed prompt instruction and provide precise prediction results along with their corresponding probabilities. The architecture of proposed MDD-LLM is shown in Figure 1. As we find that, the LLM tokenizer initially converts the input textual information into tokens that are comprehensible to the model. Subsequently, these tokens undergo feature extraction through the core architecture of the LLM. After being processed by self-attention and cross-attention mechanisms, we obtain more discriminative representations. Ultimately, the model should use these features to forecast outcomes and present the results in the requested format. Specifically, all textual components of the input prompt must be tokenized and embedded by MDD-LLM's tokenizer and word embedding layers before being fed into the decoder. The decoding process can be

outlined as follows:

$$r_t^p = f(P, r_{1:t-1}^p), \tag{1}$$

where $r_t^p \in \mathbb{V}$ is the predict token at the step $t$, and $\mathbb{V}$ defines the vocabulary set. The predicted results can be expressed as $R^P = \{r_1^p, r_2^p, ..., r_T^p\}$, and $T$ is the length of the generated diagnosis results. The language modeling loss is defined as follows:

$$\mathcal{L}_{LLM} = -\sum_{t=1}^{T} \log p(r_t^p \mid P, r_{1:t-1}^p). \tag{2}$$

This paper aims to develop a professional solution based on LLMs for MDD diagnosis. To achieve this objective, it is imperative to incorporate a significant volume of high-quality, real-world data related to depression to effectively fine-tune our proposed method. Adopting different techniques can also significantly influence the model performance in the parameter-efficient fine-tuning task for LLMs. Many different large model fine-tuning schemes have been proposed, such as LoRA [31], Q-LoRA [32], Prefix Tuning [33], and Adapter Tuning [34]. After carefully considering the task requirements and available computational resources, we select LoRA and Q-LoRA to fine-tune the proposed MDD-LLM.

## 2.4 Implementation Details

In this study, we conduct extensive experiments utilizing the Llama3.1 series models [35], with a particular emphasis on assessing the performance of the Llama3.1-8B and Llama3.1-70B models following fine-tuning. The fine-tuning process employs the LoRA method, with a rank of 8 and an alpha value of 16. The model is fine-tuned on approximately 300k pieces of data using linear warm-up and decay schedules with a peak learning rate of 0.0003, constant weight decay of 0.1, global batch size of 192, and total training epochs of 5. In the fine-tuning process of the model, the proposed method should provide answers of *Yes* or *No* based on the input information to represent the corresponding classification label, along with the associated prediction probability. Notable, the predicted probability for each class reflects the likelihood of the LLM generating its token sequence, normalized across all classes. All experimental methods are implemented using PyTorch and Python. The proposed MDD-LLM and comparison approaches are trained on four Nvidia H100 GPUs. We conduct all the experiments on the collected dataset from UK Biobank cohort. The dataset is split 80% / 20% for training / testing by patient ID. We adopted 5-fold cross-validation to verify the prediction performance of the proposed model, which can reduce the impact of partition randomness on the obtained results.

**2.5 Evaluation Metrics**

We evaluate model performance using several commonly used metrics, such as Accuracy (Acc), F1-score (F1), Specificity (SPE), Sensitivity (SNE), Positive Predictive Value (PPV), Negative Predictive Value (NPV), and Area Under the Curve (AUC). Detailed explanations of these metrics are provided in Appendix S2.

## 3. Results

**3.1 Experimental Setting**

This paper is dedicated to employing an extensive dataset of real-world samples to fine-tune existing LLMs, aiming to create a more efficient solution for MDD diagnosis. Therefore, we conduct various in-depth analyses based on the UK Biobank cohort. First, we evaluate the performance of our proposed MDD-LLM against traditional machine learning and deep learning methods using the test set. Next, we explore the factors that may influence the performance of MDD-LLM. For example, we examine different approaches to fine-tuning our proposed framework, including LoRA and Q-LoRA. Additionally, we consider how tabular data conversion methods may significantly impact the performance of LLM-based solutions. Therefore, we experiment with a range of conversion approaches, including List Template, Text Template, and GPT Generation Template. Since LLMs have not been extensively studied in MDD diagnosis, it is challenging to find LLM-based solutions to compare with our proposed method. Currently, most model-based solutions utilize machine learning or deep learning algorithms. To demonstrate the effectiveness of our proposed solution, we develop several classical methods utilizing the same dataset, such as XGBoost, Random Forest

(RF), Support Vector Machine (SVM), and Multi-Layer Perceptron (MLP).

**3.2 Model Performance on UK Biobank**

To showcase the effectiveness of MDD-LLM, we carry out comprehensive experiments on the UK Biobank dataset, followed by a comparison with existing machine learning and deep learning solutions using various evaluation metrics, such as Accuracy (ACC), F1 Score (F1), and Receiver Operating Characteristic Curve (ROC). The detail results are shown in Table 3, the ROC results are illustrated in Figure 2. The proposed MDD-LLM 8B and 70B models demonstrate superior performance when compared to existing model-based methodologies across the majority of evaluation metrics. For instance, the MDD-LLM 70B model achieves results of 0.8378, 0.8184, and 0.8919 (95% CI: 0.8799 - 0.9040) in terms of Accuracy, F1 score, and AUC, respectively. When compared to SVM, the MDD-LLM 70B shows a relative improvement of 23.57% in Accuracy and 25.87% in F1 score. While the performance of the MDD-LLM 8B solution may not rival that of the MDD-LLM 70B, it still offers a significant enhancement over existing approaches in terms of accuracy and robustness.

Table 3. Comparison results of different methods on the UK Biobank dataset.

| Method | ACC | F1 | AUC | SPE | SENS | PPV | NPV |
| --- | --- | --- | --- | --- | --- | --- | --- |
| SVM | 0.6780 | 0.6502 | 0.7450 | 0.6945 | 0.6582 | 0.6423 | 0.7091 |
| RF | 0.6868 | 0.6327 | 0.7357 | 0.7647 | 0.5933 | 0.6776 | 0.6929 |
| LightGBM | 0.7077 | 0.6691 | 0.7678 | 0.7558 | 0.6500 | 0.6893 | 0.7216 |
| XGBoost | 0.7055 | 0.6688 | 0.7481 | 0.7486 | 0.6539 | 0.6843 | 0.7218 |
| CatBoost | 0.7102 | 0.6701 | 0.7736 | 0.7627 | 0.6472 | 0.6945 | 0.7217 |
| MLP | 0.6852 | 0.6287 | 0.7507 | 0.7676 | 0.5863 | 0.6777 | 0.6901 |
| ResNet1D | 0.7062 | 0.6629 | 0.7739 | 0.7653 | 0.6354 | 0.6929 | 0.7158 |
| LLaMA3.1 8B | 0.6153 | 0.5215 | 0.6374 | 0.7436 | 0.4612 | 0.5999 | 0.6235 |
| MDD-LLM 8B | 0.7904 | 0.7627 | 0.8566 | 0.8023 | 0.7748 | 0.7509 | 0.8225 |
| MDD-LLM 70B | 0.8378 | 0.8184 | 0.8919 | 0.8358 | 0.8405 | 0.7974 | 0.8721 |

* ACC = accuracy, F1 = f1 score, AUC = area under the receiver-operating characteristic curve, SPE = specificity, SEN = sensitivity, PPV = positive predictive value, NPV = negative predictive value.

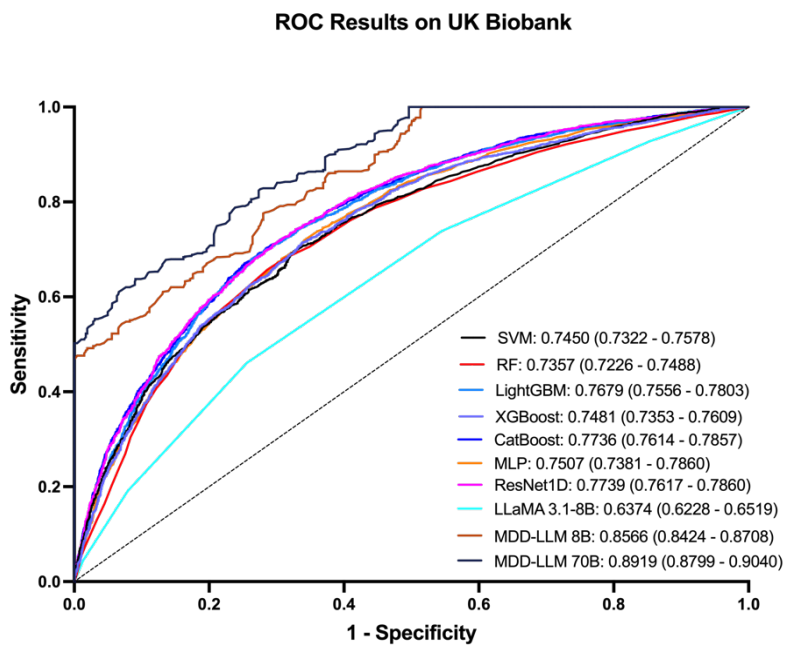

Figure 2. The comparison ROC results of different methods on UK Biobank cohorts.

### 3.3 Effectiveness of Fine-tuning Methods

Finding efficient methods to fine-tune LLMs is a valuable area for exploration. Several widely recognized techniques for model fine-tuning include LoRA, QLoRA, Adapter Tuning, Prompt Tuning, and Prefix Tuning. These methods can enhance model performance and adaptability. Given the abundance of fine-tuning methodologies available, this section concentrates on the two most prevalent techniques: LoRA and QLoRA. Notable, all of the experiments are based on MDD-LLM 8B. The comparison results are shown in Table 4. Experimental results demonstrate that the model fine-tuned by LoRA exhibits a slightly superior performance compared to the model fine-tuned with QLoRA. Notably, the LoRA solution also allows for a more efficient fine-tuning process, requiring only 40 minutes to complete training, whereas the QLoRA necessitates approximately 55 minutes. However, LoRA requires more GPU memory than QLoRA. We can find that under the same experimental settings, LoRA requires 245 GB of GPU memory for model training, whereas QLoRA requires only 172 GB. Given the substantial hardware resource demands of LLM-based solutions, we should select suitable fine-tuning techniques based on real-world scenarios.

Table 4. Model performance of MDD-LLM 8B fine-tuned by LoRA and QLoRA.

| Method | Accuracy | F1 Score | Training Times | Used Memory |
| --- | --- | --- | --- | --- |
| LoRA | 0.7904 | 0.7627 | 40 min | 245 GB |
| QLoRA | 0.7883 | 0.7624 | 55 min | 172 GB |

### 3.4 Effectiveness of Prompt Generation Types

Creating effective prompts is crucial for fine-tuning LLMs, leading to the emergence of prompt engineering tasks. In this part, we investigate how different prompt types affect model performance. We design various prompt expressions for the same tabular data, including the List Template, Text Template, and GPT Generation Template. Specifically, the List Template refers to a structured set of column names and feature values, with a fixed order for the columns. The text template expresses each feature in a sentence, such as "The patient's age is 25 years old." The GPT generation template suggests using the ChatGPT API to create a natural language description for the tabular data. Figure 3 presents an example that compares various conversion methods. Details of the tabular transformation are provided in the Appendix S1. The generated prompts are used to refine our proposed MDD-LLM 8B, and the results of these experiments are detailed in Figure 4 and the Appendix Table S2. Our findings indicate that prompts generated using the List Template yield inferior performance in fine-tuning the MDD-LLM compared to these created using the Text Template and GPT Generation Template. Notably, using prompts generated from the Text Template and GPT Generation Template yielded similar results in fine-tuning the proposed model. Given data acquisition costs, collection time, and model performance, we recommend using the Text Template for prompt generation.

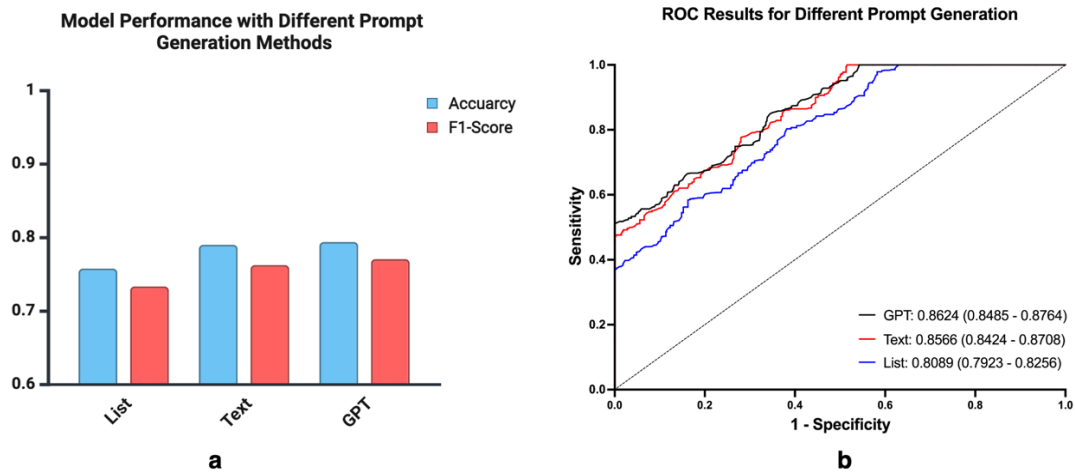

Figure 3. An example of the result of tabular data conversion of List Template, Text Template, and GPT Generation Template.

Figure 4a. Model performance with different prompt generation methods in terms of Accuracy, F1 Score. Figure 4b. The results of ROC with three different tabular data transformation approaches. The List, Text, and GPT defines transformation method of List Template, Text Template and GPT Generation Template, respectively.

## 3.5 Robustness of Feature Missing

Classic deep learning or machine learning models typically require the completeness and diversity of training samples. Therefore, numerous works have conducted extensive research on missing data and proposed several solutions, such as MICE [36] and

MissForest [37]. However, missing data is a prevalent challenge in medical-related tasks. Excessive reliance on data completion solutions may introduce considerable noise into the model, adversely affecting its performance. In contrast, LLMs possess an extensive repository of prior knowledge. For instance, they inherently understand the correlation between abnormal blood sugar levels and diabetes, as well as the potential link between prolonged working hours and the risk of depression. Research indicates that medical diagnostic solutions utilizing LLMs demonstrate impressive resilience in handling incomplete data. To investigate this issue further, we randomly retain 20%, 40%, 60%, and 80% of the features from the training data. As demonstrated in Figure 5 and Appendix Table S2, the proposed MDD-LLM exhibits strong robustness against missing data. When 60% of the features are missing, the accuracy and F1 Score of the MDD-LLM only decline slightly compared to the results from using the full dataset. In contrast, both XGBoost and MLP demonstrate significant challenges in maintaining effective performance under these circumstances.

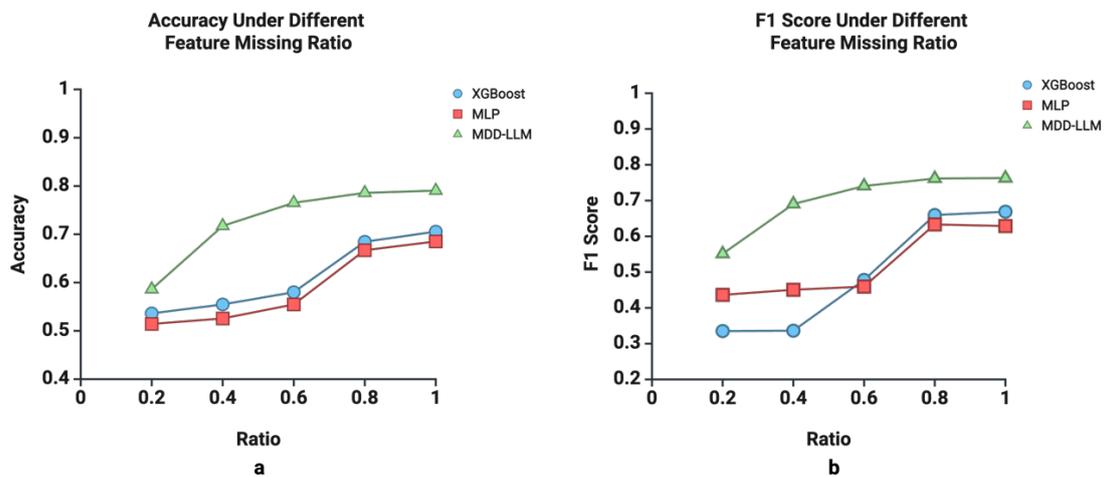

Figure 5a. Model performance with different feature retain ratio in term of Accuracy. Figure 5b. Model performance with different feature retain ratio in term of F1 Score.

## 4. Discussion

The diagnosis of MDD is often considered a complex and challenging topic within the field of mental health. Currently, the primary method for diagnosing MDD depends on scales such as the DSM-5 and PHQ-9. However, these methods require significant cooperation from the patient and a high level of experience from the clinician. Once the patient intentionally withholds information, it significantly increases the risk of erroneous conclusions and misjudgments. As technology continues to evolve, more advanced preliminary diagnostic schemes for MDD are being proposed. For instance, Wang et al. [38] introduced the M-MDD scheme, utilizing a multi-task deep learning model in conjunction with electroencephalogram (EEG) data to efficiently diagnose MDD. This approach demonstrates promising results within the validation dataset. Since MDD is related to human movements and facial expressions, numerous researchers have undertaken investigations utilizing multimodal information. For example, Sadeghi et al. [39] proposed a solution using deep learning methods to complete the MDD diagnosis task by integrating facial expression and textual information. While these models demonstrate effective performance, they necessitate a considerable quantity of high-quality data for training purposes. However, the processes of data collection and annotation present significant challenges, particularly in the context of medical-related tasks. Most model-based solutions function as black boxes, which complicates the process of elucidating the knowledge acquired by the model. The deficiency of model interpretability presents a significant barrier to applying these methods in real-world medical scenarios. The rapid advancement of LLMs has led some researchers to investigate their potential applications in diagnosing diseases [40]. For example, Kim et al. [41] used GPT4 to identify cardiovascular disease and achieved favorable results. Currently, there is a shortage of available LLMs specifically fine-tuned on extensive real-world datasets for diagnosing MDD. LLMs offer significant advantages in model accuracy, reasoning capabilities, and prior knowledge, which render them highly effective solutions for diagnosing MDD.

In light of the current challenges associated with the diagnosis of MDD and the benefits conferred by LLMs, this paper introduces MDD-LLM. This innovative approach utilizes extensive real-world samples to fine-tune the LLM, enabling a more accurate diagnosis framework for MDD. In this article, we explore various aspects of applying LLMs to MDD diagnosis. We compare the performance of MDD-LLM against traditional deep learning and machine learning methods. The experimental results show that the fine-tuned LLM-based solution achieves promising prediction accuracy and demonstrates exceptional potential in the domain of MDD diagnosis. However, fine-tuning LLMs is a complex project that still requires further research, as many factors may significantly affect the model's performance. Therefore, we conduct a series of explorations to analyze these issues. Initially, we discuss two mainstream fine-tuning technologies, LoRA and QLoRA, while examining their advantages, limitations, and suitable application scenarios. Subsequently, we review three different methods for generating prompts and determine the most suitable data conversion scheme based on usage costs and model performance.

LLM-based MDD diagnosis solutions offer notable advantages over existing model-based frameworks, including robustness to missing data and stronger model interpretability. In order to demonstrate the robustness of our proposed solution in the presence of missing data, we undertake comprehensive experiments to validate its effectiveness and compare it with several existing methods. Experimental results show that MDD-LLM maintains better performance even with substantial data absence. For instance, when 60% of the features in the training data are absent, the performance of the MDD-LLM declines by 9.24% and 9.50% in terms of Accuracy and F1 Score, respectively. In contrast, existing model-based solutions struggle to function effectively under the same circumstances. Additionally, LLM-based solutions can utilize dialogue to output prediction results and provide corresponding explanations. For instance, we transform tabular data into prompts and feed them into the MDD-LLM. The model interprets this information according to instructions and predicts whether the patient is suffering from MDD. Then, we can require the model to explain the reasoning behind its predictions. The conversational capabilities and extensive background knowledge of LLMs allow them to provide detailed explanations for their predictions. Examples illustrating the model's interpretability are shown in Figure 6.

## 5. Limitations and Future Works

In this study, we develop an MDD diagnosis solution based on LLMs and large-scale real-world samples. Experimental results indicate that the proposed MDD-LLM shows significant advantages in accuracy, robustness, and interpretability. However, several issues require further investigation. In this paper, we fine-tune the Llama 3.1 8B and 70B models utilizing the UK Biobank cohort. However, the Llama 3.1 series consists exclusively of general models, lacking specialized versions tailored for the medical field. Further exploration is needed to assess the impact of a medical LLM replacing Llama 3.1 on MDD-LLM's performance. Moreover, the interpretability of LLM-based solutions significantly surpasses that of traditional deep learning and machine learning models. Nevertheless, the hallucination in LLMs remains a challenge that requires attention. While there are standard fine-tuning schemes for LLMs, it remains unclear whether these approaches are suitable for medical-related tasks, particularly in diagnosing MDD. Due to hardware limitations, we utilize two fine-tuning methods: LoRA and QLoRA. The effectiveness of other fine-tuning approaches still requires further experimental validation.

Next, we will continue to study the LLM-based depression diagnosis model from the following aspects. First, we will evaluate various foundational LLMs to identify the most suitable option for our solution. Second, we will focus on developing LLM fine-tuning techniques specifically tailored for medical-related tasks. Third, we will investigate the hallucination issues related to the MDD-LLM and explore effective strategies to mitigate these problems. Finally, we will collect more real-world and multi-center samples for MDD diagnosis.

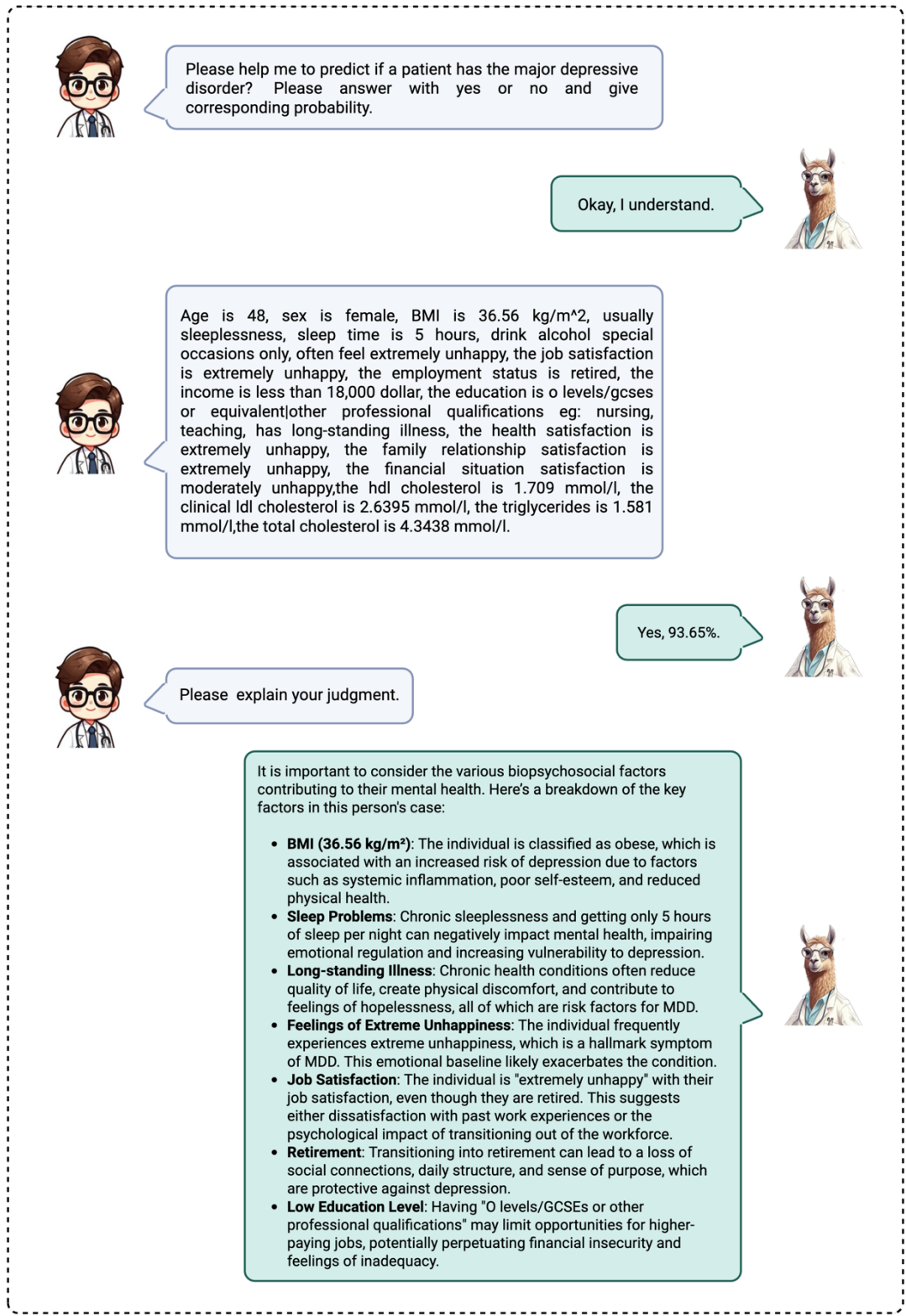

Figure 6. Example showing MDD-LLM's capability in model interpretability, reasoning, and dialogue capability.

## 6. Conclusion

In this study, we develop an MDD diagnosis solution based on LLMs and large-scale real-world samples, named MDD-LLM. To our knowledge, the proposed MDD-LLM is the first LLM-based solution fine-tuned on an extensive real-world dataset for MDD diagnosis. In order to demonstrate the model's performance, we conduct comprehensive comparisons with existing machine learning and deep learning methods. Results show that the MDD-LLM is superior to most model-based schemes across multiple evaluation metrics, including Accuracy, F1 Score, and AUC. Since few studies have explored the use of LLMs for diagnosing MDD, we discuss various factors that could influence the performance of the fine-tuned LLM-based solution. These factors include methods for transforming tabular data, the effectiveness of different fine-tuning strategies, and the model's robustness in handling missing data. Experiments have demonstrated that the MDD-LLM, fine-tuned with large-scale data, can be effectively employed for MDD diagnosis.

**Author Contribution**

Concept and design: Y. Sha, K. Li. Acquisition, analysis, or interpretation of the data: Y. Sha, H. Pan, W. Meng, G. Luo, W. Xu, Y. Du, C. Shi. Code and model build: Y. Sha, C. Shi. Drafting of the manuscript: Y. Sha, H. Pan. Critical revision of the manuscript for important intellectual content: K. Li. All the authors discussed the results and commented on the manuscript. All authors agree to be accountable for all aspects of work ensuring integrity and accuracy.

**Availability of Data and Materials**

The source code used and/or analyzed during the current study is available through GitHub (https://github.com/syysha0k/MDD-LLM) for research purpose only. This research has been conducted using the UK Biobank Resource under Application Number 99946.

**Declaration of Competing Interest**

The authors declare that they have no conflicts of interest.

**Funding**

This work was supported by the fund from Macao Polytechnic University (RP/FCA-14/2023), Macao Polytechnic University Internal Research Grant (RP/FCSD-02/2022), Science and Technology Development Funds (FDCT) of Macao (0033/2023/RIB2).

2024;14:93.

14. Blundell E, De Stavola BL, Kellock MD, Kelly Y, Lewis G, McMunn A, et al. Longitudinal pathways between childhood BMI, body dissatisfaction, and adolescent depression: an observational study using the UK Millenium Cohort Study. Lancet Psychiatry. 2024;11:47–55.

15. Darling AM, Young BE, Skow RJ, Dominguez CM, Saunders EFH, Fadel PJ, et al. Sympathetic and blood pressure reactivity in young adults with major depressive disorder. J Affect Disord. 2024;361:322–32.

16. Buckman JEJ, Saunders R, Stott J, Arundell L-L, O'Driscoll C, Davies MR, et al. Role of age, gender and marital status in prognosis for adults with depression: An individual patient data meta-analysis. Epidemiol Psychiatr Sci. 2021;30:e42.

17. Patel V, Burns JK, Dhingra M, Tarver L, Kohrt BA, Lund C. Income inequality and depression: a systematic review and meta-analysis of the association and a scoping review of mechanisms. World Psychiatry. 2018;17:76–89.

18. Touvron H, Martin L, Stone K, Albert P, Almahairi A, Babaei Y, et al. Llama 2: Open Foundation and Fine-Tuned Chat Models. 2023.

19. Bedi S, Liu Y, Orr-Ewing L, Dash D, Koyejo S, Callahan A, et al. Testing and Evaluation of Health Care Applications of Large Language Models: A Systematic Review. JAMA. 2025;333:319–28.

20. Liu Q, Wu X, Zhao X, Zhu Y, Xu D, Tian F, et al. When MOE Meets LLMs: Parameter Efficient Fine-tuning for Multi-task Medical Applications. In: Proceedings of the 47th International ACM SIGIR Conference on Research and Development in Information Retrieval. New York, NY, USA: Association for Computing Machinery; 2024. p. 1104–14.

21. Ma R, Cheng Q, Yao J, Peng Z, Yan M, Lu J, et al. Multimodal machine learning enables AI chatbot to diagnose ophthalmic diseases and provide high-quality medical responses. NPJ Digit Med. 2025;8:64.

22. Moor M, Banerjee O, Abad ZSH, Krumholz HM, Leskovec J, Topol EJ, et al. Foundation models for generalist medical artificial intelligence. Nature. 2023;616:259–65.

23. Belyaeva A, Cosentino J, Hormozdiari F, Eswaran K, Shetty S, Corrado G, et al. Multimodal LLMs for Health Grounded in Individual-Specific Data. In: Maier AK, Schnabel JA, Tiwari P, Stegle O, editors. Machine Learning for Multimodal Healthcare Data. Cham: Springer Nature Switzerland; 2024. p. 86–102.

24. Sha Y, Meng W, Luo G, Zhai X, Tong HHY, Wang Y, et al. MetDIT: transforming and analyzing clinical metabolomics data with convolutional neural networks. Anal Chem. 2024;96:2949–57.

25. Sha Y, Zhai X, Li J, Meng W, Tong HH, Li K. A novel lightweight deep learning fall